\documentclass{article}
\usepackage{spconf,amsmath,graphicx}
\usepackage{hyperref} 
\usepackage{booktabs}
\usepackage{amssymb,xcolor,stackengine,graphicx}
\usepackage{multirow}
\usepackage{textpos}

\title{On the Utility of Self-supervised Models for Prosody-related Tasks}

\name{%
\begin{tabular}{@{}c@{}}
Guan-Ting Lin$^{1 \dagger}$, Chi-Luen Feng$^{1 \dagger}$, Wei-Ping Huang$^{1 \ddagger}$, Yuan Tseng$^{1 \ddagger}$, Tzu-Han Lin$^{1 \ddagger}$, Chen-An Li$^{1 \ddagger}$,\\  \thanks{$^{\dagger \ddagger}$Equal contribution}  Hung-yi Lee$^1$, Nigel G. Ward$^2$
\end{tabular}}

\address{$^{1}$National Taiwan University, Taiwan\\$^{2}$University of Texas at El Paso, USA}

\begin{document}
%
                   
\maketitle
\begin{abstract}
Self-Supervised Learning (SSL) from speech data has produced models
that have achieved remarkable performance in many tasks, and that are known
to implicitly represent many aspects of information latently present
in speech signals. However, relatively little is known about the
suitability of such models for prosody-related tasks or the extent to
which they encode prosodic information. We present a new evaluation
framework, ``SUPERB-prosody,'' consisting of three prosody-related
downstream tasks and two pseudo tasks. We find that 13 of the 15 SSL models 
outperformed the baseline on all the prosody-related tasks. We also
show good performance on two pseudo tasks: prosody reconstruction and future prosody prediction. We further analyze the layerwise contributions of the SSL models. Overall we conclude that SSL speech models are highly
effective for prosody-related tasks. We release our code\footnote{https://github.com/JSALT-2022-SSL/superb-prosody} for the community to support further investigation of SSL models' utility for prosody.
\end{abstract}


\begin{textblock*}{\textwidth}(0cm,10cm)
\tiny
\noindent
{Copyright 2022 IEEE. Published in the 2022 IEEE Spoken Language Technology Workshop (SLT) (SLT 2022), scheduled for 19-22 January 2023 in Doha, Qatar. Personal use of this material is permitted. However, permission to reprint/republish this material for advertising or promotional purposes or for creating new collective works for resale or redistribution to servers or lists, or to reuse any copyrighted component of this work in other works, must be obtained from the IEEE. Contact: Manager, Copyrights and Permissions / IEEE Service Center / 445 Hoes Lane / P.O. Box 1331 / Piscataway, NJ 08855-1331, USA. Telephone: + Intl. 908-562-3966.}
\end{textblock*}

\section{Introduction}
\label{sec:intro}
Self-supervised Learning (SSL) has revolutionized research in many
areas of artificial intelligence, including speech processing. SSL
pre-trained speech models have shown remarkable performance and
generalizability across a wide range of tasks \cite{SUPERB, SUPERB-sg,mohamed2022self}. However, we do not currently have a good understanding of what knowledge these models capture nor of the
limits of their power. This is true in particular for the
prosodic aspects of speech.

The speech signal contains not only lexical but prosodic information.
Broadly speaking, the latter has three realms of function:
paralinguistic, phonological, and pragmatic. Paralinguistic
functions, such as marking speaker identity and expressing emotion,
are largely conveyed by prosodic settings that are stable over the span of many utterances, and are often evident from any sample
of just a few syllables. Phonological functions, notably marking
the identity of syllables and words with tones and stress patterns,
are largely conveyed by prosodic features whose temporal occurrence is
tightly linked to the units they mark. The utilities of SSL models for
these two realms have been demonstrated, by many recent studies using
tasks from the SUPERB \cite{SUPERB} benchmark, among others.

However, for the third realm, the realm of pragmatic function, the
question of the utility of SSL models has remained open. Functions in
this realm, include managing turn-taking, marking topic structure and
information structure, and expressing engagement, stance, attitude,
and intent. These pragmatic functions are especially important in
dialog, and we expect that future dialog systems will need more
prosodic competence, in order to enable more satisfying user
experiences and to support interaction in novel genres and situations
\cite{marge2022}.  In many cases, these functions are expressed using
multistream temporal configurations of low-level prosodic features,
where these configurations can last from a few hundred milliseconds to
several seconds \cite{ward-book,niebuhr-stepped}, and may be only
loosely aligned with the lexical content. As these configurations are
fundamentally different from the forms of prosody in the other two
realms, it is an open question whether SSL models are also useful for this realm.

Accordingly our research question is whether pre-trained models have
utility for prosody-conveyed pragmatic functions. We investigate this
in four ways. First, we assemble a set of prosody-intensive tasks and
measure how well pre-trained models support them. Second, we use pitch
and energy reconstruction pseudo-tasks to measure how well these
models represent prosodic information. Third, we evaluate the utility
of these models for the prediction of future pitch and energy. Fourth, we
probe the pre-trained models to see in which layers prosodic
information is likely represented.

The main contributions of this paper are: 1) The finding that
pre-trained SSL models indeed can provide value for prosody-intensive
tasks, often reaching state-of-the-art performance. 2) 
Results for 15 recent SSL models span different model architectures and  
pre-training objectives. 3) Analysis of the representation
of prosody in SSL models, including layerwise analysis. 4) An
open-source evaluation framework, SUPERB-prosody, examines the
prosodic prowess of SSL models.

\section{RELATED WORK}
\label{sec:relatedwork}
The most directly relevant study \cite{dammp} aimed primarily to
evaluate a model for producing de-identified representations of
speech, but includes three aspects that are very relevant to our
research question. First, for evaluation purposes, several ``spoken
language understanding" tasks were selected, of which three of these
were both pragmatics-relevant and prosody-intensive. Second, six
pre-trained models were tested against this task set, showing various
levels of performance. Third, probing their own model, VQP, provided
evidence that it was encoding, to some extent, several prosodic
features. Taken together these results suggest that the answer to our
research question is yes, but the case is not settled for two
reasons. First, neither their performance results nor their probing
results were compared against non-pre-trained baselines, leaving open
the question of whether the pre-trained models were in fact providing
any benefit. Second, many aspects of their methods are unclear, and
no code is available to enable replication. Thus, we need further
investigation, and an open and transparent evaluation framework for
SSL models. Very recent work has found that SSL models are helpful for
predicting some perceptions of speaking style, but require additional
downstream sequence-modeling layers for best performance
\cite{aguirre-ward-avila}.

Evaluation benchmarks have been critical in supporting and evaluating
the rise of SSL in the speech field. NOSS \cite{NOSS} is a benchmark
for non-semantic downstream tasks. SUPERB \cite{SUPERB} broadly
examines SSL models for content, speaker, semantic and paralinguistic
aspects, demonstrating that SSL models generalize across diverse
downstream tasks. SUPERB-sg \cite{SUPERB-sg} enhances the SUPERB
benchmark with more challenging semantic and generative tasks. SLUE
\cite{SLUE}, another recent benchmark, targets spoken language
understanding tasks. However, up to now, the speech community lacks an
evaluation benchmark/framework to measure the prosodic utility of SSL
models.

Analysis of speech SSL models has recently received significant
attention, with previous works examining various
aspects \cite{pasad2021layer, shah2021all, similarity_analysis,
  dammp}. \cite{pasad2021layer, similarity_analysis} mainly focus on
analyzing lexical information in SSL models. \cite{shah2021all}
investigate acoustic, syntactic, and semantic characteristics, but
they only experiment on two SSL models
and probe only by utterance-wise
regression tasks. Accordingly, more focused analysis is needed,
especially regarding prosodic information.
\vspace{-0.3cm}
\begin{figure}[t]
  \centering
  \includegraphics[width=1\linewidth]{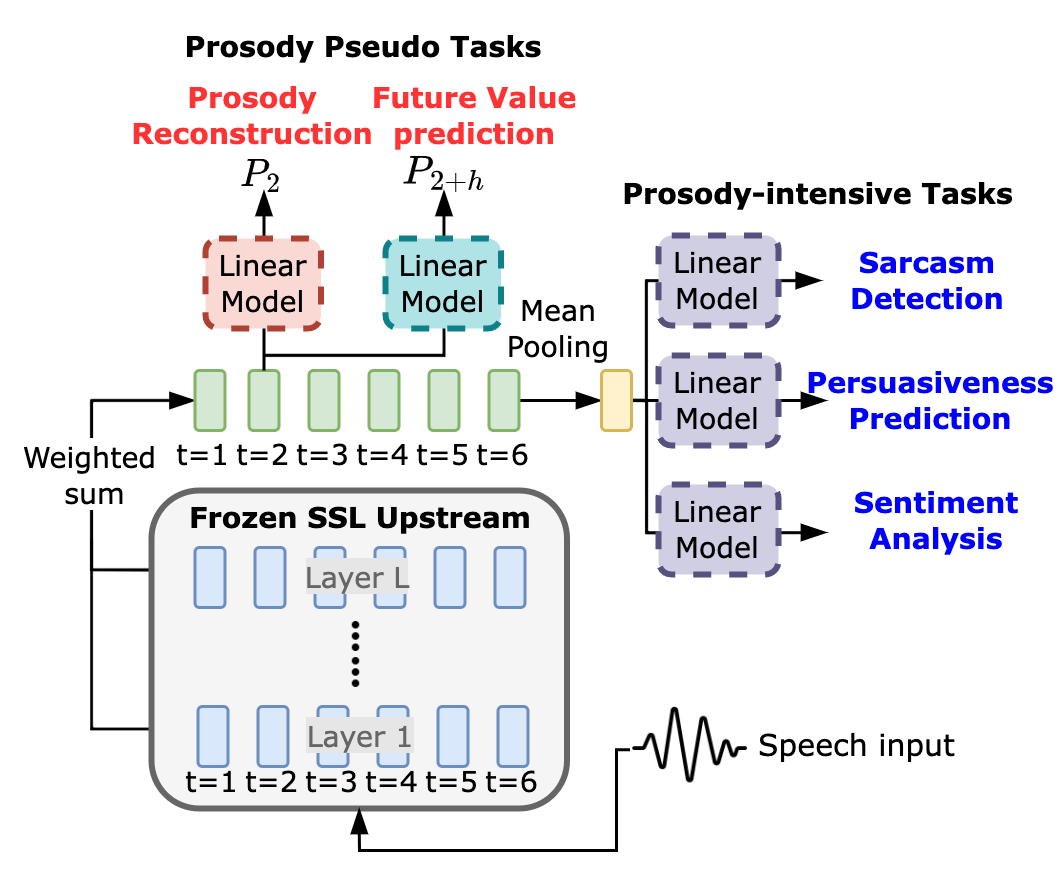}
  \vspace{-0.9cm}
  \caption{Diagram of \texttt{SUPERB-Prosody} framework. We extract
    the hidden representations from a frozen SSL model, and 
    lightweight linear models are used for each downstream task. $P_i$ means
    the value of the rule-based prosodic feature at time frame $i$. }
  \label{fig:superb-prosody}
\end{figure}

\begin{table*}[ht]
\centering
\setlength{\tabcolsep}{3pt}
\resizebox{0.75\textwidth}{!}{
\begin{tabular}{l|c|r|c|c|c|c}
\toprule
Model & Network & \#Params & Stride & Input & Corpus & Pretraining  \\ \midrule \midrule

FBANK & - & 0 & 10ms & waveform & - & - \\ \midrule

APC~\cite{apc1} & 3-GRU & 4.11M & 10ms & FBANK & LS 360 hr & F-G  \\

VQ-APC~\cite{vq_apc} & 3-GRU & 4.63M & 10ms & FBANK & LS 360 hr & F-G + VQ  \\

NPC~\cite{npc} & 4-Conv, 4-Masked Conv & 19.38M & 10ms & FBANK & LS 360 hr & M-G + VQ  \\

Mockingjay~\cite{mockingjay} & 12-Trans & 85.12M & 10ms & FBANK & LS 360 hr & time M-G  \\

TERA~\cite{tera} & 3-Trans & 21.33M & 10ms & FBANK & LS 960 hr & time/freq M-G  \\ 

\midrule

modified CPC~\cite{modified_cpc} & 5-Conv, 1-LSTM & 1.84M & 10ms & waveform & LL 60k hr & F-C \\

wav2vec~\cite{wav2vec} & 19-Conv & 32.54M & 10ms & waveform & LS 960 hr & F-C  \\

vq-wav2vec~\cite{vq_wav2vec} & 20-Conv & 34.15M & 10ms & waveform & LS 960 hr & F-C + VQ \\

DistilHuBERT~\cite{distilhubert} & 7-Conv 2-Trans & 23.49M & 20ms & waveform & LS 960 hr &  KD  \\

wav2vec 2.0 Base~\cite{wav2vec2} & 7-Conv 12-Trans & 95.04M & 20ms & waveform & LS 960 hr & M-C + VQ  \\

wav2vec 2.0 Large~\cite{wav2vec2} & 7-Conv 24-Trans & 317.38M & 20ms & waveform & LL 60k hr & M-C + VQ  \\

HuBERT Base~\cite{hubert} & 7-Conv 12-Trans & 94.68M & 20ms & waveform & LS 960 hr & M-P + VQ  \\

HuBERT Large~\cite{hubert} & 7-Conv 24-Trans & 316.61M & 20ms & waveform & LL 60k hr & M-P + VQ  \\

WavLM Base~\cite{wavlm} & 7-Conv 12-Trans & 94.68M & 20ms & waveform & LL 60k hr & M-P + VQ  \\

WavLM Large~\cite{wavlm} & 7-Conv 24-Trans & 316.62M & 20ms & waveform & Mix 94k hr & M-P + VQ  \\
\bottomrule
\end{tabular}}
\caption{SSL models examined. \#Params includes parameters for both
  pre-training and inference. LS = LibriSpeech and LS = LibriLight. For the
  pre-training methods, VQ = vector quantization, F = future, M = masked, G = generation, C = contrastive
  discrimination, P = token prediction/classification, and KD = 
  knowledge distillation. } 
\label{table:upstreams}
\end{table*}

\section{The SUPERB-Prosody Framework}
\vspace{-0.1cm}
This section introduces our tasks ---  classification, prosody reconstruction, and future value prediction task --- and our evaluation framework, including the upstream/downstream setup. 

\subsection{Classification tasks}
\vspace{-0.2cm}
To evaluate the pragmatic and prosody-related abilities of pretrained
models, we need a set of ``prosody-intensive'' tasks, that is, tasks
where it is known that prosodic abilities are useful. 
We chose three well-curated, open-source, utterance-classification
datasets, involving sentiment, sarcasm, and persuasiveness. To briefly
describe each task:

\textbf{Sentiment Analysis (SA)} involves detecting the degree of
positive or negative feeling in an utterance. While sentiment and
emotion are often conflated, and similar methods may work for both
tasks, sentiment is less visceral and of greater practical importance.
Specifically we chose the CMU-MOSEI~\cite{CMU-MOSEI} corpus, in which
each utterance is labeled from $-3$ to $3$, representing the degree of
sentiment. Following previous works, we experiment with two settings:
binary classification, with the dataset split by the labels in $[-3,
  0)$ and $(0,3]$, and seven-category classification. The evaluation
metric is accuracy.

\textbf{Sarcasm Detection (SarD)} is perhaps the most obviously prosody-intensive task, as a mismatch between the lexical content and the prosodic
message is frequently the major marker of sarcasm. We chose
the MUStARD~\cite{mustard} corpus, in which each utterance has a label of $0$ or $1$, for sarcasm or non-sarcasm. We follow the speaker-dependent setup in the original paper~\cite{mustard}, using five-fold cross-validation for evaluation. The evaluation metric is the F1 score.

\textbf{Persuasiveness Prediction (PP)} is the task of detecting whether a presentation is likely to be convincing to others. Correlates include how pleasant the speaker's voice is and their perceived confidence and dominance, all of which involve prosody. We chose the POM \cite{pom} corpus. The labels are 0 or 1, for persuasive or non-persuasive. The evaluation metric is accuracy.

\subsection{Prosody reconstruction}
Prosody Reconstruction (ProP) is a pseudo-task designed to test
whether SSL models embed specific prosody features in their hidden
representation. As our target, we chose the two most commonly used
prosody features, pitch and energy. Given the SSL features, we use a
lightweight linear downstream model to predict each. The pitch is
represented in log scale, using as targets the values computed by
pYAAPT\footnote{\href{http://bjbschmitt.github.io/AMFM\_decompy/pYAAPT.html}{http://bjbschmitt.github.io/AMFM\_decompy/pYAAPT.html}}.
For energy, we use the librosa
toolkit\footnote{\href{https://github.com/librosa/librosa}{https://github.com/librosa/librosa}},
again using a log scale. As data, we use LibriTTS~\cite{LibriTTS},
a multi-speaker text-to-speech dataset, and for both features, the evaluation
metric is the Mean Square Error (MSE) of the differences. No loss is
computed for unvoiced frames, that is, frames where pYAAPT detects no
pitch.

\subsection{Future value prediction}
Future Value Prediction (FVP) is designed to test whether the output
from SSL models can predict future prosody. The task setting is
similar to the prosody reconstruction task, using pitch and energy
features as the prediction targets and using LibriTTS dataset with MSE objective. The task is, given the information from $t=1$ to the current frame
$i$, to predict the value at $i+h$, where $h$ is the prediction
horizon. Four prediction horizons $h$, namely $0.12$, $0.24$, $0.50$,
and $1.00$ seconds, are used. Because most SSL speech models are not
causal (due to the inclusion of either self-attention or
bi-directional connections), we only test causal SSL models or
attention-based SSL models for which we can apply an attention mask to
avoid cheating with future information\footnote{We exclude WavLM for
this task because it uses a gated relative position bias, so simply
modifying the attention mask would cause a model mismatch between
training and inference.}. The evaluation metrics are MSE. 

\begin{figure*}[t]
  \centering
  \includegraphics[scale=0.195]{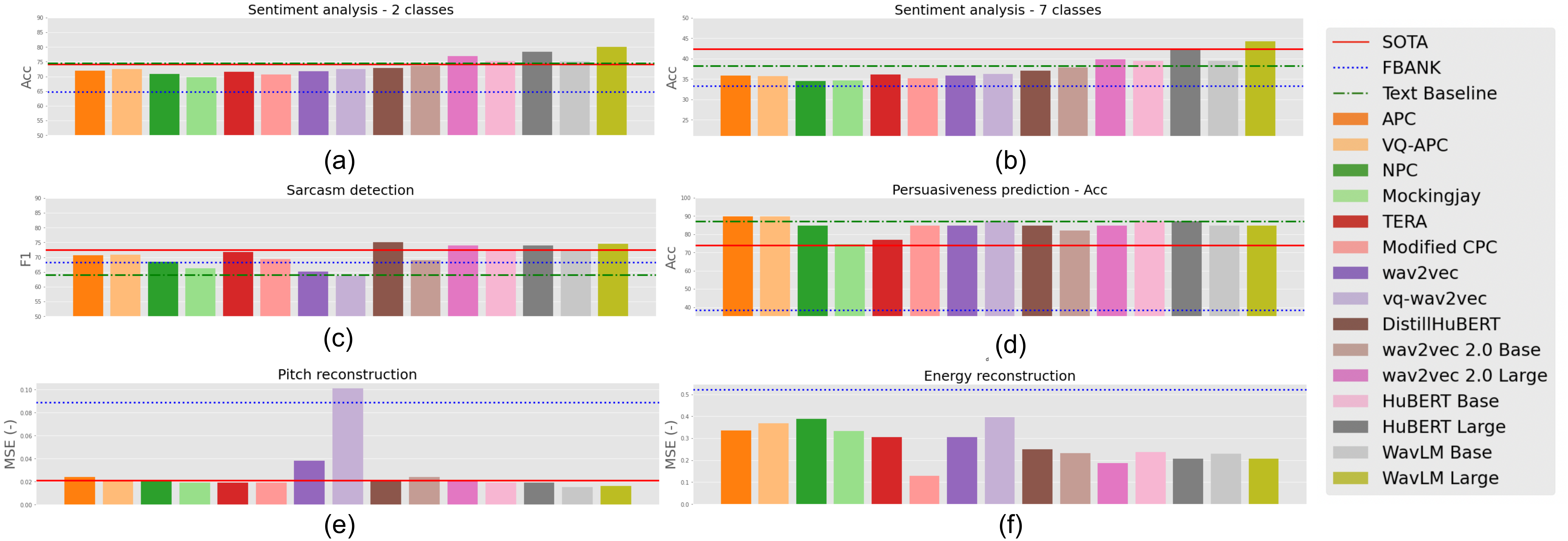}
  \vspace{-0.5cm}
  \caption{Results for Sentiment Analysis (SA),     Sarcasm Detection (SarD), Persuasiveness Prediction (PP), and Prosody Reconstruction (PR). State-of-the-art (SOTA) values are from \cite{CMU-MOSEI} for
    SA, \cite{SD-sota} for SarD, \cite{pom-SOTA} for PP, and
    REAPER\cite{reaper} for pitch reconstruction. Text-only baselines are only for the
    prosody-intensive downstream tasks, and thus not available for PR.}
  \label{fig:result}
\end{figure*}

\subsection{Evaluation framework}
\label{subsec:exp}
As suggested by Figure \ref{fig:superb-prosody}, our framework
consists of (1) an upstream SSL speech model and (2) a linear
downstream model for probing. Following the procedure of
\texttt{SUPERB} \cite{SUPERB}, the parameters of the upstream models
are fixed for all the downstream tasks.  For each frame of the input, 
we extract the representations $\mathbf{x_i}$ from each hidden layer $i$ of the
upstream model, and we aggregate those hidden representations per
utterance into $\mathbf{y} = \sum_i^L w_i \mathbf{x_i}$ where the $w_i$ are trained per task.  
The resulting $\mathbf{y}$, a two-dimensional matrix (time $\times$ aggregated features), 
is the input for the downstream model.



\subsubsection{Upstream models: Speech SSL models}
The SSL models tested are summarized in Table \ref{table:upstreams}.
These are a diverse collection, including modified CPC
\cite{modified_cpc}, APC \cite{apc1}, VQ-APC \cite{vq_apc}, NPC
\cite{npc}, TERA \cite{tera}, vq-wav2vec \cite{vq_wav2vec},
DistilHuBERT \cite{distilhubert}, HuBERT \cite{hubert}, wav2vec
\cite{wav2vec}, wav2vec 2.0 \cite{wav2vec2}, and WavLM
\cite{wavlm}. The selected SSL models span different network
architectures and pre-training objectives.

\subsubsection{Downstream models: linear probing model}
\label{downstream}

For the classification tasks, SA, SarD, and PP, the representation
$\mathbf{y}$ is mean-pooled along the time axis, forming a dense
vector of dimension (time ×
aggregated features) to (aggregated features).
This vector is then fed to a simple linear model to project
from model dimension to 1. The training objective is Cross-Entropy Minimization.

In ProR and FVP, the goal is to predict the
fine-grained prosodic information. We use the \textit{frame-level representation} from each time step of $\mathbf{y}$ as the
input, and the downstream model is a linear model, projecting from
model dimension to 1. MSE minimization is the training objective. We try multiple learning rates for each task ($[1e^{-2}$,
  $1e^{-3}$, $1e^{-4}$, $1e^{-5}$, $1e^{-6}]$), and report the best
performance. The training step is 3000 for SarD and 50000 for the
other tasks.
\vspace{-0.3cm}
\subsubsection{Baselines}
As a baseline feature set, we use``FBANK," the 80-dimensional Log Mel Filterbank features with delta and delta-delta features (240 dimensions in total), chosen because this is known to work well for many speech tasks. For the classification tasks, FBANK features are average-pooled across each utterance. 
For pitch reconstruction, we also compare the performance of another high-quality off-the-shelf pitch extractor, Talkin's REAPER. For future value prediction (FVP) we design a baseline, FBANK + RNN,
which feeds filterbank features from $t=1$ to $i$ into a one-layer uni-directional Recurrent Neural Network (RNN) with 128 hidden size to predict the value at
$t=i+h$.  

As an additional point of comparison, we also explore the text-only
performance for each classification task. Since all datasets contain
ground truth speech transcription, we take these transcriptions as the
input data. As the NLP model we use the pre-trained RoBERTa~\cite{roberta}
for SarD and SA, and Longformer~\cite{longformer} for
PP\footnote{Longformer is based on RoBERTa, but can accept up to 4096
tokens (versus 512 tokens for RoBERTa), and is used here because the
transcriptions of PP are too long for RoBERTa.}. We follow the typical
method to fine-tune the pre-trained NLP model: extract the sentence
embedding from the [CLS] token's embedding, and feed it to a
linear classification model. No parameters are frozen during
fine-tuning. We vary the learning rate ($[1e^{-3}$, $1e^{-4}$,
  $1e^{-5}$, and $1e^{-6}]$) and report the best performance.
  
\begin{figure*}[t]
  \centering
  \includegraphics[scale=0.48]{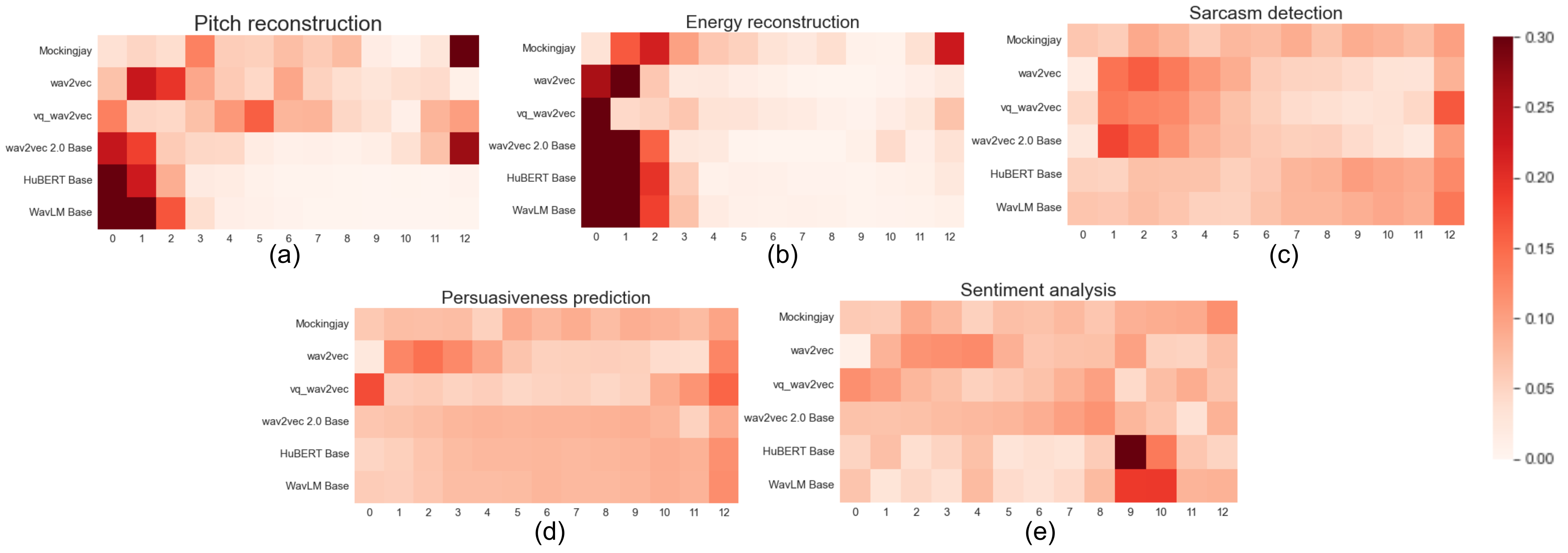}
  \vspace{-0.5cm}
  \caption{The contribution analysis for each task. The darker the color, the higher the contribution. We only include models which have 12 layers of representations.}
  \label{fig:weight_line}
\end{figure*}

\begin{table*}[ht]
\centering
\resizebox{0.66\textwidth}{!}{
\begin{tabular}{c|cccc|rrrr}
\toprule
\multirow{2}{*}{Model} & \multicolumn{4}{c|}{Pitch w/ Prediction Horizon (s)$\downarrow$} & \multicolumn{4}{c}{Energy w/ Prediction Horizon (s)$\downarrow$} \\ \cline{2-9} 
 & ~ 0.12   & ~ 0.24   & ~ 0.50    & \multicolumn{1}{c|}{~ 1.00}  & ~~ 0.12        & ~~ 0.24    & ~~ 0.50   & ~~ 1.00        \\ \midrule \midrule
FBANK + RNN             &  0.049      &  0.104      &  0.142      &      0.157                     &    0.52    &    1.42     &   2.06     &    2.46    \\
\midrule
APC & 0.033 & 0.043 & 0.052 & 0.053          & 0.91 & 1.46 & 1.98 & 2.37 \\
modified CPC & 0.038 & 0.051 & 0.062 & 0.065 & 0.79 & 1.37 & 1.94 & 2.47 \\
wav2vec & 0.053 & 0.064 & 0.075 & 0.075      & 0.68 & 1.15 & 1.70 & 2.31 \\
Mockingjay              &  0.069      & 0.077       &  0.081      &      0.099              &    0.48         &       1.92  &    2.17   &   2.43    \\
wav2vec 2.0 Base        & 0.038  & 0.047  & 0.047  & 0.049                     & 0.44       & 0.85       & 1.24       & 1.49      \\
wav2vec 2.0 Large       & 0.035  & 0.039  & 0.045  & 0.046                     & 0.43       & 0.80       & 1.23       & \textbf{1.31}     \\
HuBERT Base             & 0.029  & 0.036  & 0.041  & 0.042                     & \textbf{0.39}       & \textbf{0.70}       & \textbf{1.12}  & 1.42      \\
HuBERT Large            & \textbf{0.025}  & \textbf{0.027}  & \textbf{0.028}  & \textbf{0.037}           & 0.41       & 0.77       & 1.21   & 1.36      \\
\bottomrule
\end{tabular}}
\vspace{-0.2cm}
\caption{The MSE loss for FVP. Lower values indicate better prediction, of future pitch or energy.}
\label{table:FVP-results}
\end{table*}


\begin{table*}[ht]
\centering
\resizebox{0.62\textwidth}{!}{
\begin{tabular}{l cc cc cc}
\toprule
 & \multicolumn{2}{c}{SarD F1$\uparrow$} & \multicolumn{2}{c}{PP Acc$\uparrow$} & \multicolumn{2}{c}{SA Acc$\uparrow$}\\
\cmidrule(lr){2-3} \cmidrule(lr){4-5} \cmidrule(lr){6-7}
Layer selection     &  (0,1,\underline{12})  & (10,11,\underline{12}) &  (0,1,\underline{12})  & (10,11,\underline{12}) &  (0,1,\underline{8})  & (7,\underline{8},9)  \\
\midrule
wav2vec 2.0 Base & \textbf{70.6} & 66.3 & \textbf{81.3} & 81.3  & 73.6 & \textbf{74.0}   \\
HuBERT Base      & \textbf{72.0}& 70.0 & \textbf{85.8}  & 83.8 & 75.2  & \textbf{76.2}  \\
\bottomrule
\end{tabular}
}
\vspace{-0.2cm}
\caption{Experimental results for layer-limited feature integration for SarD, PP, and SA. The best layer, underlined, is chosen based on the contribution analysis. }
\label{table:integration}
\end{table*}


\begin{table}[ht]
\centering
\resizebox{0.4\textwidth}{!}{
\begin{tabular}{l|llll}
\toprule
Method           & ZH-p$\downarrow$  & \multicolumn{1}{l}{PO-p$\downarrow$}  & ZH-e$\downarrow$   & \multicolumn{1}{l}{PO-e$\downarrow$}\\ \midrule
FBANK       & 0.050 & 0.096                     & 0.849 & 0.477  \\
vq-wav2vec       & 0.022 & 0.097                     & 0.490 & 0.339  \\
wav2vec 2.0 Base & 0.012 & 0.039                     & 0.338 & 0.160  \\
HuBERT Base      & 0.009 & 0.042                     & \textbf{0.232} & \textbf{0.149}   \\
WavLM Base       & \textbf{0.008} & \textbf{0.019}                     & 0.233 & 0.192   \\
\bottomrule
\end{tabular}}
\vspace{-0.2cm}
\caption{MSE for prosody reconstruction, showing cross-lingual transferability. 
p = pitch, e = energy.}
\label{tab:cross-lingual}
\end{table}

\vspace{-0.3cm}
\section{Main Results}
\subsection{SSL models perform well on prosody-intensive tasks}
\label{subsec:pi-tasks}
The experimental results for SA, SarD, and PP are shown in Figure
\ref{fig:result}. In Figure \ref{fig:result} (a) and (b), we can see
that all SSL models yield better SA performance than the baseline
FBANK features in both 2-label and 7-label evaluations. The large SSL
models, wav2vec 2.0, HuBERT, and WavLM, even improve on the
state-of-the-art (SOTA) performance \cite{CMU-MOSEI} in the 2-label
setup. In the 7-label setup, WavLM Large outperforms audio-only SOTA
performance. Around one-third of the SSL models show better
performance than the text-only baseline, confirming the value of
acoustic information for SA.


Figure \ref{fig:result} (c) shows the SarD result. Although
Mockingjay, wav2vec, and vq-wav2vec show inferior performance to the
baseline FBANK, the other SSL models do well, with DistilHuBERT,
HuBERT, and WavLM improving on the previous audio-only SOTA
\cite{SD-sota}. In SarD, all acoustic models, both SSL
models and FBANK, outperform the text-only baseline, confirming that
SarD requires acoustic information beyond content for prediction.


Lastly, for the PP results, Figure \ref{fig:result} (d), shows that all
SSL models yield better performance than the FBANK features and the
previous audio-only SOTA \cite{pom-SOTA}. APC, VQ-APC, vq-wav2vec, and
HuBERT obtains superior or equal accuracy to the text-only baseline.

\subsection{SSL models encode prosodic information}

While the results above suggest that the SSL models are encoding
prosodic information, there is also direct evidence from the ProR and
FVP tasks, as seen in Figure \ref{fig:result} and Table
\ref{table:FVP-results}.  \looseness=-1

For PR, the two features, pitch and energy, have slightly different
results. In pitch reconstruction, Figure \ref{fig:result} (e), all the
SSL models perform better than baseline FBANK except for
vq-wav2vec. Several SSL models surpass REAPER performance, with the
WavLM the best. As for energy reconstruction, Figure \ref{fig:result}
(f) shows that all SSL models greatly improved over baseline
FBANK. Although generation-based SSL models perform well on pitch reconstruction, they did relatively
worse on energy reconstruction. On the other hand, the SSL models
pre-trained by masked contrastive discrimination/token prediction show
strong performance on both pitch and energy reconstruction. Overall,
we observe that SSL models indeed encode prosodic information.

FVP is more challenging than ProR since it requires the model to
capture both global and local prosodic information for successful future 
prediction. From Table \ref{table:FVP-results}, we see,
unsurprisingly, that the larger prediction horizons make prediction
harder. For pitch, HuBERT Large gets the best performance,
outperforming other SSL models and baseline FBANK+RNN at all four
horizons. Although the pre-training objectives of APC, modified CPC,
and wav2vec  involve future generation/discrimination, they
still result in inferior performance to wav2vec 2.0 and HuBERT. As for
future energy prediction, only wav2vec 2.0 and HuBERT consistently
outperform baseline FBANK.
In general, we observe that some SSL models are not good at FVP, but wav2vec 2.0 and HuBERT outperform FBANK by a large
margin. This result suggests that wav2vec 2.0 and HuBERT do have the
capability to encode and summarize relevant prosodic information.

\section{Further Analysis}
\label{sec:analysis}

\subsection{Layerwise contribution analysis}

In order to estimate the contribution of each layer, we consider two factors.
First, we use the weight $w_i$ from each layer (through the weighted-sum mechanism) to the
downstream model. Second, because typical feature magnitudes may vary,
we also consider the values of the hidden representation $\mathbf{x}$ in each layer, 
as measured by the L2-norm of feature values for the testing data. After getting the whole
L2-norm features for each layer for each sample, we take the mean
across samples to get the feature magnitude estimation. The contribution
for each layer is then defined as: $c_i = ||\mathbf{x_i}||_2 \times w_i$, where $i$ means
the layer number.


The results are shown in Figure~\ref{fig:weight_line}. 
From Figure \ref{fig:weight_line}(a), we can see that for most SSL models, 
the contribution is strongest in the first few layers for both pitch and energy reconstruction. 
This shows that SSL tends to best represent prosodic information in the front. 
One exception is Mockingjay, where the largest contribution is located in the last layer. 
Because Mockingjay's pre-training objective is to reconstruct frame-wise features, 
it is unsurprising that the later layers contain a good representation of low-level prosody. 

However, for the classification tasks SarD and PP, 
Figure \ref{fig:weight_line} (c) and (d) shows that the distribution of layer contributions is smooth, 
suggesting that both tasks need information across multiple layers. 
As previous work has shown that later layers represent more content information ~\cite{pasad2021layer, wavlm}, 
this suggests that both prosodic and content information are needed for SarD and PP. 
As for SA, in Figure \ref{fig:weight_line}(e), we observe a high contribution value in the latter layers for the high-performing models
HuBERT and WavLM. This suggests
 that SA might only require content information to perform well.


\subsection{Feature integration}
\label{subsub:integration}

To further examine whether the encodings in the first few layers
bring the most benefit, we designed a new experiment for three
prosody-intensive tasks using wav2vec 2.0 and HuBERT. We compare two
settings, concatenation of 1) the features from the first two layers
and the best layer we discovered\footnote{The best layer is determined
by the contribution analysis above.}, and 2) the best layer with its two neighbor
layers. These concatenated features are passed to the downstream
model. The downstream model size of the two settings is the same, so
we can compare the performance fairly. If the first setting is better
than the second setting, this would indicate that early-layer (low-level)
information indeed most benefits the final results.  \looseness=-1

The results are shown in Table~\ref{table:integration}. 
In SarD and PP, we see the integration of the first two layers yields better performance, which means the low-level information improves the modeling of SarD and PP. Yet the use of low-level information does not improve SA performance, 
suggesting
that SA may rely on content rather than just low-level information. 


\subsection{Cross-lingual transferability}
Further, we did a preliminary investigation of whether SSL models may
have cross-lingual transferability for prosodic information, using the
ProR task as in previous experiments: specifically, attempting pitch
and energy reconstruction.  While the pitch of one frame is primarily
a physical phenomenon, in context there may be language dependencies.
All the SSL models being pre-trained in English, we try this for
Mandarin and Polish, using data from AISHELL-3 \cite{AISHELL-3_2020}
multilingual LibriSpeech (MLS) \cite{MLS}, respectively. 
Table \ref{tab:cross-lingual}
summarizes the results.   We note that WavLM
is good at pitch reconstruction, and HuBERT is superior at energy
reconstruction.



\section{Conclusion}
\label{sec:conclusion}
\vspace{-0.2cm}
In this work, we explore the utility of SSL models for
prosody-conveyed pragmatic functions. Using the newly-proposed
SUPERB-prosody evaluation framework, we find that SSL models provide
significant value for prosody-intensive tasks, and that they are good at extracting prosodic information in 
pseudo tasks. Furthermore, we analyze the layer contribution and
discover that most SSL models tend to store prosodic information in the
first few layers. 
However, the field still lacks a good
understanding of why different SSL models are better for different
tasks \cite{dumpala22_interspeech}, and this is an important topic for future work. 

\section{Acknowledgement}
We thank the Taiwan Web Service and the National Center for High-performance Computing (NCHC) of the National Applied Research Laboratories (NARLabs) in Taiwan for providing computing and storage resources. Part of the work presented here was carried out during the 2022 Jelinek Memorial Summer Workshop on Speech and Language Technologies at Johns Hopkins University, which was supported with unrestricted gifts from Amazon, Microsoft, and Google.

\bibliographystyle{IEEEbib}
\bibliography{strings,refs}

\end{document}